\documentclass[11pt,a4paper]{article}
\usepackage{acl2019}
\usepackage{stfloats}
\usepackage{latexsym}
\usepackage[T2A]{fontenc} 
\usepackage[utf8]{inputenc}
\usepackage[english,russian]{babel}
\usepackage{tempora} % {times} for Russian
\usepackage{enumitem}
\usepackage{url}
\usepackage{multirow}
\usepackage{graphicx}
\usepackage[normalem]{ulem}
\usepackage{ulem}
\usepackage{gb4e}
\noautomath

\aclfinalcopy % Uncomment this line for the final submission

\title{AGRR-2019: A Corpus for Gapping Resolution in Russian}

\author{First Author \\
  Affiliation / Address line 1 \\
  Affiliation / Address line 2 \\
  Affiliation / Address line 3 \\
  \texttt{email@domain} \\\And
  Second Author \\
  Affiliation / Address line 1 \\
  Affiliation / Address line 2 \\
  Affiliation / Address line 3 \\
  \texttt{email@domain} \\}

\author{
Maria Ponomareva$^{\spadesuit}$~~~
Kira Droganova$^{\dag}$~~~
Ivan Smurov$^{\spadesuit, \clubsuit}$~~~
Tatiana Shavrina$^{\diamondsuit, \heartsuit}$\\
$^{\spadesuit}$ABBYY, Moscow, Russia \\[1mm]
$^{\dag}$Charles University, Faculty of Mathematics and Physics \\[1mm]
$^{\clubsuit}$Moscow Institute of Physics and Technology, Moscow\\[1mm]
$^{\diamondsuit}$Sberbank, Moscow, Russia \\[1mm]
$^{\heartsuit}$National Research University Higher School of Economics, Moscow\\[1mm]
{\tt \{maria.ponomareva,ivan.smurov\}@abbyy.com}\\
{\tt droganova@ufal.mff.cuni.cz}\\
{\tt Shavrina.T.O@sberbank.ru}\\
}

\date{}

\begin{document}
\maketitle
\begin{abstract}
This paper provides a comprehensive overview of the gapping dataset for Russian that consists of 7.5k sentences with gapping (as well as 15k relevant negative sentences) and comprises data from various genres: news, fiction, social media and technical texts. The dataset was prepared for the Automatic Gapping Resolution Shared Task for Russian (AGRR-2019) - a competition aimed at stimulating the development of NLP tools and methods for processing of ellipsis. \\

In this paper, we pay special attention to the gapping resolution methods that were introduced within the shared task as well as an alternative test set that illustrates that our  corpus is a diverse and representative subset of Russian language gapping sufficient  for effective utilization of machine learning techniques.
\end{abstract}

\section{Introduction}

During the last two years gapping (i.e., the omission of a repeated predicate which can be understood from context ~\cite{ross70}) has received considerable attention in NLP works, both dedicated to parsing ~\cite{schuster2018sentences, kummerfeld2017parsing} and to corpora enhancement and enrichment~\cite{nivre2018enhancing, biblio:DrGiMindGap2018}. At the same time, just a few works dealt with compiling a corpus that would represent different types of ellipsis, and almost exclusively for English. Most of these works address VP-ellipsis, which refers to the omission of a verb phrase whose meaning can be reconstructed from the context~\cite{johnson2001vp}, for instance, in “Mary loves flowers. John does too”~\cite{hardt1997empirical, nielsen2005corpus, bos2011annotated}. The research has mainly been conducted so far on rather small amounts of data, not exceeding several hundreds of sentences. In this work we aim to create a resource with a decent amount of data that would include a broad variety of genres and would rely minimally on any specific NLP frameworks and parsing systems.

This work consists of four parts. First, we describe the dataset, its features, and provide examples of Russian-specific constructions with gapping. Second, we describe an alternative test set that we have prepared to demonstrate that our corpus is representative enough. Then we briefly describe the key metrics that have been proposed to evaluate the quality of gapping resolution methods within the shared task. Finally, we provide a detailed analysis of the methods that have successfully solved the gapping resolution task as well as the results that were achieved on the alternative test.

\section{Gapping}
\label{Gapping}
We confine ourselves to the types of elliptical constructions for Russian that involve omission of a verb, a verb phrase or a full clause.

In this work we use the following terminology for gapping elements. We call the pronounced elements of the gapped clause remnants. Parallel elements found in a full clause that are similar to remnants both semantically and syntactically are called remnant correlates. The missing material is called the gap ~\cite{coppock2001gapping}.

Traditionally, gapping is defined as the omission of a repeating predicate in non-initial composed and subordinate clauses where both remnants to the left and to the right remain expressed.

\begin{exe}
\ex
\gll Я принял её за итальянку, а его за шведа.\\
  I mistook her for Italian and him for Swede\\
\trans \small {‘I mistook her for Italian and \sout{I mistook} him for Swede’}
\end{exe}

However, a broader interpretation is possible ~\cite{Testelets:2011}. Some features of gapping worth mentioning are listed below.

Elements remaining after predicate omission can be of different types. Consider the following examples where remnants are predicates (2), preposition phrases (3), adverbs (4), adjectives (5) potentially with their dependents.

\begin{exe}
\ex
\gll Одно может вдохновлять, а другое вгонять в тоску.\\
 one can inspire and other put in melancholy\\
\trans \small {‘One thing can inspire and the other \sout{can} put you in a melancholic mood.’}
\end{exe}

\begin{exe}

\ex
\gll Советую вам поменьше думать о проблемах, и побольше — об их решении.\\
 recommend you less think about problems and more - about their solution  \\
\trans \small{‘I recommend you to think less about problems, and \sout{think} more about their solutions.’}
\end{exe}

\begin{exe}
\ex
\gll Вначале они играли интересно, потом – прескучно.\\
 at.first they played interesting.ADV after - boring.ADV.INT\\
\trans \small{‘At first they played interestingly, then \sout{they played} extremely dully.’}
\end{exe}

\begin{exe}
\ex
\gll Сердце ее было слишком чистым, чувства слишком искренними.\\
 heart her was too pure feelings  too sincere\\
\trans \small{‘Her heart was too pure and her feelings \sout{were} too sincere.’}
\end{exe}

The set of constructions for Russian that implement stripping ~\cite{Merchant:2016} seems to be broader than for English and the difference between gapping and stripping in Russian is less clear. We encountered a wide variety of examples that go beyond the canonical examples. Examples (6) and (7) illustrate the cases when arguments/adjuncts of  the elided verb do not fully correspond to the arguments/adjuncts of the pronounced verb, thus some of them (\textit{в конце} ‘in the end’ in (6), \textit{за 2009 год} ‘during year 2009’) do not have correlates. We consider such examples gapping with one remnant and include them in the corpus.
\begin{exe}
\setlength\itemsep{0em}
\ex
\gll  Добавляем муку, крахмал и разрыхлитель, а \textbf{в} \textbf{конце} сметану.\\
 add flour starch and baking.powder and in end sour.cream\\
\trans \small{‘We add flour, starch and baking powder, and at the end \sout{we add} sour cream.’}
\end{exe}
\begin{exe}
\ex
\gll Рост	цен	составил	11,9	процента	(\textbf{за} \textbf{2009} \textbf{год}	-	4,4	процента)
\\
 growth	prices	amounted.to	11.9	percent	in	2009	year -	4.4	percent\\
\trans \small{‘Price growth amounted to 11.9 percent (in 2009  \sout{it amounted to} 4.4 percent)’}
\end{exe}

\section{Corpus Description}

Since the publicly available markup with gapping is sparse, one of our key motivations was to create a corpus that contains as many examples of gapping as possible. To the best of our knowledge, no other publicly available dataset contains a comparable amount of gapping examples. 

With that in mind, we decided to base our corpus on the markup obtained with Compreno~\cite{anisimovich2012}. Compreno is a syntactic and semantic parser that contains a module for predicting null elements in the syntactic structure of a sentence. An overview of the module can be found in~\cite{bogdanov2012}.

While cleaning up the output of a specific system allows us to obtain markup much faster than annotating from scratch, training on the resulting corpus may yield systems that would reproduce the original system’s output instead of properly modeling the real-world natural language phenomenon. We took this risk because even if the corpus we have created contains Compreno bias, the selection is representative enough. Moreover, in order to further test for the presence of such bias, we evaluated the top systems of the shared task on an alternative test set that was created from SynTagRus (see Section \ref{SynTagRus}).

The corpus is available on the shared task's GitHub \footnote{https://github.com/dialogue-evaluation/AGRR-2019}.
\subsection{Annotation Scheme}
We utilize the following labels for fully annotated sentences with gapping:
\begin{itemize}[topsep=0pt,itemsep=0pt,partopsep=0pt, parsep=2pt]
 \setlength\itemsep{0em}
\item The gap is labeled $V$.
\item The head of the pronounced predicate corresponding to the elided predicate is labeled $cV$. 
\item  Remnants and their correlates are labeled $Rn$ and $cRn$ respectively, where n is the pair’s index
\end{itemize}

For gapping annotation we use square brackets to mark all gapping elements (whole NP, VP, PP etc. for remnants and their correlates and the predicate controlling the gap), the gap is marked with \textit{[\textsubscript{V}]}. Example (8) shows an example of bracket annotation of (1).
\begin{exe}
\ex
\gll {Я [\textsubscript{cV}} {принял] [\textsubscript{cR1}} {её] [\textsubscript{cR2}} за итальянку], {а [\textsubscript{R1}} {его] [\textsubscript{R2}} за шведа].\\
  I mistook her for Italian and him for Swede\\
\trans \small{‘I mistook her for Italian and \sout{I mistook} him for Swede’}
\end{exe}

Therefore, the full list of annotation labels is as follows: $cV$, $cR1$, $cR2$, $V$, $R1$, $R2$.

\subsection{Obtaining the Data}
In this section we provide a detailed description of the process of compiling the corpus.
The bulk of the collection comprises Russian texts of various genres: news, fiction, technical texts. To our understanding, many NLP tasks that could benefit from gapping resolution are often applied to social network data. Therefore, we balanced the corpus by adding texts from the popular Russian social network VKontakte. They make up a quater of the collection.

First, all texts in the text collection were parsed with Compreno.
We identified the sentences in which gapping was predicted. Using the Compreno parser, we generated bracketed annotation for each sentence (in which every gapping element $X$ has an opening bracket $[X$ and closing bracket $]$).

Mindful of our main goal (i.e., to maximize the amount of data in the corpus), we decided to avoid fixing the annotation errors manually. Instead 11 assessors were asked to evaluate the annotation, assigning one of four classes:
\begin{description}[topsep=0pt,itemsep=0pt,partopsep=0pt,parsep=2pt]
 \setlength\itemsep{0em}
\item[0] no gapping, no markup is needed; 
\item[1] all gapping elements are annotated correctly;
\item[2] some gapping elements are annotated incorrectly;
\item[3] problematic example.
\end{description}

Each sentence was evaluated by two assessors. Table \ref{table:assessment} shows that  41\% out of 17411 sentences have correct annotation and 19\% were erroneously attributed to the examples with gapping, according to both annotators.

\begin{table}[htbp]

\begin{center}
\resizebox{1.0\columnwidth}{!}{%
\begin{tabular}{|l|l|l|l|l|}
\hline

\textbf{} & \textbf{0} & \textbf{1} & \textbf{2} & \textbf{3} \\ \hline
\textbf{0} & \textbf{3350 (19\%)} & 370 (2.1\%) & 303 (1.7\%) & 254 (1.5\%) \\ \hline
\textbf{1} & 394(2.3\%) & \textbf{7201(41\%)} & 1163 (6.7\%) & 283 (1.6\%) \\ \hline
\textbf{2} & 288 (1.7\%) & 581 (3.3\%) & 1960 (11\%) & 302 (1.7 \%) \\ \hline
\textbf{3} & 446 (2.5 \%) & 230 (1.3\%) & 153 (0.9\%) & 133 (0.8 \%)\\ \hline

\end{tabular}%
}
\end{center}
\caption{Assessment analysis for the AGRR corpus; 0, 1, 2, 3 - annotation classes.}
\label{table:assessment}
\end{table}

The main application of our corpus is in machine learning, therefore the corpus has to include negative examples (i.e., sentences without gapping). We considered two types of negative examples to select more relevant sentences. The first type comprises problematic negative sentences on which the Compreno parser false positively predicted gapping (labeled 0 by both assessors). Introducing negative examples of this type (i.e. hard negatives) supposedly would allow a system to improve upon the results of the source parser. The second type comprises sentences of at least 6 words that contain a dash or a comma, and a verb. We made the negative class twice as large as the positive one.

It is worth mentioning that cases marked 2 and 3 noticeably overlap with cases of gapping from the SynTagRus gapping test set, which we use to validate our AGRR corpus (see section \ref{SynTagRus}; for cases 2 and 3 examples see the official shared task report ~\cite{AGRR-2019}).

The test set contains ten times fewer examples than the combined training and development sets with the same distribution of genres - 75\% from fiction and technical literature, 25\% from social media -  and the same 1:2 ratio of positive to negative classes.

\begin{table}[h!]
\begin{center}
\resizebox{1.0\columnwidth}{!}{%
\begin{tabular}{|l|l|r|r|r|r|r|}

\hline
\multicolumn{2}{|c|}{}  & \multicolumn{2}{|c|}{0} & \multicolumn{2}{|c|}{1}  & \multicolumn{1}{|c|}{ sum }            \\ \hline
\multirow{2}{*}{$\mathbf{dev}$}& vk   & 670   & \multirow{2}{*}{2760} & 326  &  \multirow{2}{*}{1382} &   \multirow{4}{*}{20548} \\ \cline{2-3}
                               & other & 2090  &                      & 1056 &                        &                          \\ \cline{1-6}
\multirow{2}{*}{$\mathbf{train}$}& vk & 2860 & \multirow{2}{*}{10864} & 1366 & \multirow{2}{*}{5542} &                           \\ \cline{2-3}
                                & other & 8004 &                      & 4176 &                       &                           \\ \cline{1-7}
\multirow{2}{*}{$\mathbf{test}$}& vk & 343 & \multirow{2}{*}{1365} & 185 & \multirow{2}{*}{680} & \multirow{2}{*}{2045} \\ \cline{2-3}
                                & other & 1022 &  & 495 &  &  \\ \hline
\multicolumn{2}{|c|}{$\mathbf{sum}$} &  & 14989 &  & 7604 & $\mathbf{22593}$ \\
\hline
\end{tabular}%
}
\end{center}%
\caption{\# examples by class; vk stands for social media texts}%
\label{table:data}%
\end{table}%
\subsection{Dataset Format}
When choosing the annotation format, we aimed to minimize reliance on any specific NLP frameworks and parsers. Since tokenization is often an integral part of NLP pipelines, we decided not to provide any gold standard tokenization and thus did not choose the commonly used CoNLL-U format.

Instead, markup of each sentence contains a class label (1 if gapping is present in the sentence, 0 otherwise) and character offsets for each gapping element (no offsets if sentence does not contain the corresponding gapping element).

\section{SynTagRus Gapping Test Set}
\label{SynTagRus}

In order to test how well our corpus represents the phenomenon in question, we employ an alternative test set\footnote{https://lindat.mff.cuni.cz/repository/xmlui/handle/11234/1-3001} obtained from SynTagRus - the dependency treebank for Russian that provides comprehensive manually-corrected morphological and syntactic annotation~\cite{bog2009, dyc2015}.

To detect and extract relevant sentences, we rely on the original SynTagRus annotation~\cite{iomdin2009structure}, i.e., the Nodetype attribute, which, if present with the value “FANTOM”, indicates an omission in surface representation.

All the sentences were manually verified and divided into three categories: 
\begin{description}[topsep=0pt,itemsep=0pt,partopsep=0pt, parsep=2pt]
\item [1]  cases similar to the ones encountered in the AGRR corpus;
\item [2]  cases of gapping not included in the AGRR corpus;
\item [3]  cases considered other types of ellipsis rather than gapping.
\end{description}

Sentences from all three categories as well as the number of aooripriate negative examples (obtained from SynTagRus with simple heuristics) will be 
further jointly referred to as the SynTagRus gapping test set.

We expect the systems trained on the AGRR corpus to show better results for category 1, because the examples may differ stylistically and thematically but not on a structural level. High scores obtained for category 2 would demonstrate that the corpus and the top systems were transferable to a broader range of gapping cases. Additionally, we provide the results obtained by the top systems for category 3.

We further illustrate the diversity of ellipsis cases in categories 2 and 3 using examples adapted from the SynTagRus gapping test corpus.

\subsection{Gapping not Included in the AGRR Corpus}
In Russian, the number of remnants is limited only by the valency of the predicate and can exceed two. Consider an example (9) with three remnants. 
\begin{exe}
\ex
\gll [\textsubscript{cR1} В {Испании] [\textsubscript{cR2}} в 1923 {году] [\textsubscript{cV}} установил] {диктатуру [\textsubscript{cR3}} {генерал Педро де Ривера], [\textsubscript{R1}} в {Польше] [\textsubscript{R2}} в 1926-м] {- [\textsubscript{R3}} Пилсудски].\\
. In Spain in 1923 year established  dictatorship {general Pedro de Rivera}, in Poland in 1926 -  Pilsudski \\
\trans \small{‘In Spain, the dictatorship of General Pedro de Rivera was established in 1923, while in Poland the dictatorship \sout{was established} by  Pilsudski in 1926.’}
\end{exe}

The AGRR corpus does not contain examples where the order of remnants differs from the order of correlates, though the structure is possible under certain conditions~\cite{Paducheva:74}.

\begin{exe}
\ex
\gll [\textsubscript{cR1} Школа и {уроки] [\textsubscript{cV}} {принадлежали] [\textsubscript{cR2}} кругу мучительных обязанностей], {а [\textsubscript{R2}} душевному выбору] {-  [\textsubscript{R1}} зеленая птица с красной головой]. \\
 . school and lessons belonged.to circle painful duties and soul.ADJ choice - green bird with red head\\
\trans \small{‘School and lessons belonged to the circle of painful duties, while a green bird with a red head \sout{belonged} to the choice of the soul.’}
\end{exe}

The cases with two independent instances of gapping are not seen by the systems trained on the AGRR corpus. In (11) the bracketed sentence has its own gapping with overt predicate имеет ‘has’ not connected to the first occurrence of gapping, where predicate достигает ‘reaches’ is elided.

\begin{exe}
\ex
\gll [\textsubscript{cR1} Ширина {долины] [\textsubscript{cV}} {\textbf{достигает}] [\textsubscript{cR2}} 600 {км], [\textsubscript{R1}} глубина] {- [\textsubscript{R2}} 8 км] (для сравнения: {Большой каньон [\textsubscript{cV}} {\textbf{имеет}] [\textsubscript{R1}} {ширину] [\textsubscript{R2}} до 25 {км] [\textsubscript{R1}} и {глубину] [\textsubscript{R2}} 1,8 км]). \\
. width valley reaches 600 km, depth - 8 km for comparison {Grand Canyon} has width to 25 km and  depth 1.8 km   \\
\trans \small{‘The width of the valley reaches 600 kilometers, the depth \sout{reaches} 8 kilometers (for comparison: the width of the Grand Canyon is about 25 kilometers and the depth \sout{is} 1.8 kilometers).’}
\end{exe}

In Russian, gapping is not necessarily formed by omission of a verb. See (12), where the elided predicate is a noun (отчуджение ‘isolation’).

\begin{exe}
\ex
\gll Бюрократизм привел {к [\textsubscript{cV}} {отчуждению] [\textsubscript{cR1}} {трудящихся] [\textsubscript{cR2}} от {власти], [\textsubscript{R1}} {крестьян] [\textsubscript{R2}} от земли]. \\
red.tape led to alienation working.people from power peasants from  land   \\
\trans \small{‘Red tape led to the alienation of working people from power, and \sout{alienation} of peasants from the land’}
\end{exe}

The SynTagRus gapping test set contains several examples illustrating a particular type of gapping that we refer to as gapping with generalization. In this type of gapping, the correlate clause semantically generalizes over instances described in subsequent gapped clauses. Furthermore, the main clause may lack the correlates of some remnants, e.g. промышленностью ‘industry’,  наукой ‘science’ in (13).
\begin{exe}
\ex
\gll [\textsubscript{cR1} Средства и способы] создаются талантливыми учеными, {а [\textsubscript{cV}} {реализуются]: [\textsubscript{R1}} средства] {- [\textsubscript{R2}} \textbf{военной} \textbf{промышленностью}], {а [\textsubscript{R1}} способы] {- [\textsubscript{R2}} \textbf{военной} \textbf{наукой} \textbf{и} \textbf{опытом}] \\
  . means and methods are.created talented scientists and are.realized means - military industry and methods - military science and experience. \\
\trans \small{‘Means and methods are created by talented scientists, and are realized: the means \sout{are realized} by the military industry, and the methods \sout{are realized} by military science and experience.’}
\end{exe}

According to~\cite{Kazenin2007}, gapping in Russian cannot elide an intermediate node in the tree structure. However, our data shows that such elision is possible. Consider (14), where the left correlate is higher syntactically than the elided predicate.

\begin{exe}
\ex
\gll {Если [\textsubscript{cR1}} {можно] [\textsubscript{cV}} {передать] [\textsubscript{cR2}} один университет], то {почему [\textsubscript{R1}} {нельзя]  [\textsubscript{R2}} другие]?!  \\
 if is.possible transfer.INF one university, then why not.possible others \\
\trans \small{‘If it is possible to transfer one university, then why can't others \sout{be transferred}?!’}
\end{exe}
\subsection{Other Types of Ellipsis}

Along with cases of gapping not included in the AGRR corpus, we categorized sentences from the SynTagRus gapping test set that contain types of ellipsis other than gapping. Below we provide frequent categories of ellipsis with illustrations.

 Ellipsis in comparative constructions ~\cite{Bacs:2018, KenMerch:2000} has restrictions that differ from gapping. 
 
\begin{exe}
\ex
\gll От сна за рулем погибает {столько же} водителей, сколько от алкоголя   \\
 from sleeping behind wheel die as.many  drivers how.many/as from alcohol \\
\trans \small{‘As many drivers die from sleeping behind the wheel, as \sout{many drivers die} from alcohol’}
\end{exe}

Cases where the second remnant is missing and the second clause contains just one remnant are called stripping ~\cite{Merchant:2016}. Canonical examples of stripping are limited to a small number of constructions (16) - (17). According to~\cite{hankamersag1976}, who introduced the term: “Stripping is a rule that deletes everything in a clause under identity with corresponding parts of a preceding clause except for one constituent (and sometimes a clause-initial adverb or negative).” 
\begin{exe}
\ex The man stole the car after midnight, \textbf{but not} the diamonds. ~\cite{Merchant:2016}
\ex  Abby can speak passable Dutch, and Ben, \textbf{too}. ~\cite{topless}
\end{exe}

Our SynTagRus gapping test corpus contains examples with more (нет in (18)) and less canonical (причем in (19)) markers, but all of them can be distinguished  from gapping with one remnant by the presence of closed set markers (see Section \ref{Gapping}).

\begin{exe}
\ex
\gll Тогда деньги стали общими, а экономики – \textbf{нет}.    \\
Then money became shared and economy - not.  \\
\trans \small{‘Then the money became shared, but the economy did not \sout{become shared}.’}
\end{exe}

\begin{exe}
\ex
\gll  В Сталинграде каждый сражается, \textbf{причем} как   мужчины, {так и} женщины  \\
 in Stalingrad, everyone fights and both men and women.’\\
\trans \small{‘In Stalingrad, everyone continuously fights, both men and women \sout{fight}.’}
\end{exe}

Another type of ellipsis encountered in the SynTagRus gapping test corpus is sluicing~\cite{Merch:2001}. Sluicing deletes the predicate from an embedded interrogative clause with no arguments remaining.

\begin{exe}
\ex
\gll Медикам дается указание как-то бороться с этим явлением, а как – никому не известно.   \\
doctors are.given instructions somehow cope with this phenomenon and how - no.one NEG knows  \\
\trans\small{‘Doctors are instructed to somehow cope with this phenomenon, but no one knows how \sout{to cope with it}.’}
\end{exe}

Finally, in the SynTagRus gapping test set there are numerous sentences with the following type of ellipsis: the repeating predicate is elided leaving only its arguments, and there are no correlates for arguments in the full clause. In sentences of this category, the second clause adds further details to the situation mentioned in the full clause.

Consider (20), where the predicate меняются (‘they change’) has no subject in the full clause, while it is added in the elided clause with одним игроком (‘by one player’).

\begin{exe}
\ex
\gll Правила меняются по ходу игры и всегда почему-то одним игроком \\
 rules are.changed with progress game and always for.some.reason one player.INST \\
\trans \small{‘The rules are changed as the game progresses and for some reason \sout{the rules are changed} always by one player’}
\end{exe}

In (22) the elided clause adds the manner справкой(‘by certificate’) to the action подтвердить (‘to verify’) 

\begin{exe}
\ex
\gll Студент должен подтвердить свои доходы, причем желательно справкой.  \\
 student must confirm his income and preferably  certificate.INST \\
\trans \small{‘The student must confirm their income, and preferably \sout{confirm} with a certificate.’}
\end{exe}

\section{Shared Task}

In this paper, we revisit the information about the shared task that is essential for understanding the results of this paper (for details see the shared task report~\cite{AGRR-2019})

We have formulated 3 different tasks concerning gapping with increasing complexity:
\begin{enumerate}[topsep=0pt,itemsep=0pt,partopsep=0pt, parsep=2pt]
 \setlength\itemsep{0em}
\item Binary presence-absence classification - for every sentence, decide if there is a gapping construction present.
\item Gap resolution - for every sentence with gapping, predict the position of the elided predicate and the head of the pronounced predicate in the antecedent clause.
\item Full annotation - for every sentence with gapping, predict the linear position of the elided predicate and positions of its remnants in the clause with the gap, as well as the positions of remnant correlates and the head of the pronounced predicate in the antecedent clause.
\end{enumerate}

Solutions of all three tasks can be utilized by researchers studying gapping. Since sentences with gapping are naturally rare, the solution of the binary classification task will help researchers to find sentences with gapping for further analysis and data enrichment. Solutions of the other two tasks can be used to facilitate gapping resolution for parsing systems as well as to verify the quality of gapping annotation in syntactic corpora.
\subsection{Metrics}
The main metric for the binary classification task is standard f-measure.
Two other tasks were scored based on symbolwise f-measure on gapping elements relevant to the particular task (all 6 for full annotation, V and cV for gap resolution). 

The following is a description of symbolwise f-measure:
\begin{itemize}[topsep=0pt,itemsep=0pt,partopsep=0pt, parsep=2pt]
\item true negative samples for binary classification task do not affect total f-measure;
\item for true positive samples, symbolwise f-measure is obtained for each relevant gapping element separately, thus generating 6 scores for the full annotation task and 2 scores for the gap resolution task (if the evaluated sentence is either false positive or false negative, all the generated scores are equal to 0);
\item the obtained f-measures are macro-averaged over the whole corpus.
\end{itemize}

One particular feature of the described metrics is that the second and the third task scores cannot exceed the first task score and thus binary classification errors are relatively harshly penalized in all three tasks. We have deliberately chosen such metrics since ellipsis is a rare language phenomenon and thus misclassification (false positive in particular) should be treated with caution.

\section{Results and Analysis}
\subsection{Evaluation Results}
\label{Evaluation}

Results of the top two participants on both the AGRR-2019 and the SynTagRus gapping test set are presented in Table \ref{table:scores}. The implemented solutions are described in detail in the next section.
The full table with shared task results as well as brief description of each participating system is available in the official report.

\begin{table}[htbp]
\begin{center}
\resizebox{1.0\columnwidth}{!}{%
\begin{tabular}{|l|l|r|r|r|}
\hline
\textbf{Corpus} & \textbf{Team} & \textbf{Binary} & \textbf{Gap} & \textbf{Full} \\ \hline
\multirow{2}{*}{AGRR} & Winner & 0.96 & 0.90 & 0.89 \\ \cline{2-5}
 & 2nd best   & 0.95 & 0.86 & 0.84 \\ \hline
\multirow{2}{*}{SynTagRus} & Winner & 0.91 & 0.76 & 0.77 \\ \cline{2-5}
 & 2nd best & 0.88 & 0.67 & 0.64 \\ \hline
\end{tabular}%
}
\end{center}
\caption{Top systems F1 scores on AGRR-2019 and SynTagRus test set. Binary: binary classification; Gap: gap resolution; Full: full annotation.}
\label{table:scores}
\end{table}

F1 scores on the SynTagRus gapping test set are measured for the subset consisting of categories 0 and 1. While examples of categories 2 and 3 cannot be reliably measured with the shared task metrics, we have calculated the number of examples of each category classified by the top systems as gapping. These results are shown in Table \ref{table:fps}.

\begin{table}[htbp]
\begin{center}
\resizebox{1.0\columnwidth}{!}{%
\begin{tabular}{|l|l|l|r|r|}
\hline
\textbf{Cat} & \textbf{\begin{tabular}[c]{@{}l@{}}Total \\ \end{tabular}} & \textbf{Team} & \textbf{\begin{tabular}[c]{@{}l@{}} positives\end{tabular}} & \textbf{\begin{tabular}[c]{@{}l@{}} positives\end{tabular}, \%} \\ \hline
\multirow{2}{*}{0} & \multirow{2}{*}{1166} & Winner & 8 & 0.7\% \\ \cline{3-5}
 & & 2nd best & 30 & 2.6\% \\ \hline
\multirow{2}{*}{1} & \multirow{2}{*}{507} & Winner & 433 & 85.4\% \\ \cline{3-5}
 & & 2nd best & 420 & 82.8\% \\ \hline
\multirow{2}{*}{2} & \multirow{2}{*}{75} & Winner & 26 & 35\% \\ \cline{3-5}
 & & 2nd best & 37 & 49\% \\ \hline
\multirow{2}{*}{3} & \multirow{2}{*}{100} & Winner & 6 & 6\% \\ \cline{3-5}
 & & 2nd best & 13 & 13\% \\ \hline
\end{tabular}%
}
\end{center}
\caption{Number of sentences classified as gapping for each category of SynTagRus gapping test set.}
\label{table:fps}
\end{table}
Table \ref{table:scores} demonstrates that the AGRR-2019 corpus contains enough data for effective utilization of machine learning techniques. The results on the SynTagRus gapping test set in particular show that systems trained on the AGRR-2019 corpus are able to yield reasonably good results on a dataset obtained without any usage of the Compreno parser. While both systems experience a performance drop relative to scores on the AGRR-2019 test set, this can be attributed to domain shift (as two corpora have different genre composition etc.). In our opinion these results provide enough evidence to state that while the AGRR-2019 corpus has some inherent restrictions (see Section \ref{SynTagRus}), it reflects a real-world linguistic phenomenon rather than the output of the Compreno system.

Performance on category 0 examples, as is shown in Table \ref{table:fps}, demonstrates that high-precision systems can be trained on the AGRR-2019 corpus\footnote{It can be argued that the second best system has high false positive rate relative to the frequency of gapping in natural language. However one should keep in mind that classes 0 and 1 had 2:1 distribution in the training set. Changing this balance in favour of negative examples may potentially increase the precision of the systems.
Moreover, manual analysis of these false positives shows that some of these examples do in fact contain gapping while many others are borderline.}.

Performance on category 2 examples demonstrates that such systems can potentially recognize gapping examples of types completely unrepresented in the training set (obviously, performance on such sentences could be improved if similar examples were be added to the training set).

Performance on category 3 examples, by contrast, demonstrates that such systems can differentiate gapping from other types of ellipsis (including rather similar ones such as stripping and sluicing).

\subsection{General Analysis}

Most participants, including all top systems, treated gap resolution and full annotation tasks as sequence labeling tasks. The most popular approaches were to enhance the standard BLSTM-CRF architecture~\cite{lample2016, ma&hovy2016}, to pretrain an LSTM-based language model or to use transformer-based solutions ~\cite{vaswani2017attention, bert}.
 
Most participating systems did not use any token-level features other than word embeddings, character-level embeddings, or language model embeddings ~\cite{elmo, bert, ulmfit}. Of particular note is that neither of the 2 top-scoring systems used morphological or syntactic features. While it may be theorized that using such features could yield some improvements, we presume that language model embeddings (especially when coupled with self-attention as in the top two systems) contain most syntactic information relevant to ellipsis resolution.
 
\subsection{Top Systems Analysis}

The top two systems share several important elements: language model embeddings, self-attention (the winner as part of BERT, the second best team solution directly), and the part of the system designed to choose sound label chains (FSA-based postprocessor for the winner, NCRF++ for the second best team; ~\cite{ncrf}). 
The third element is necessary when solving the task as sequence labeling (and more task-specific FSA-postprocessing yields better results). We can assume the first two elements combined contain most syntactic and semantic information relevant to ellipsis resolution.

The top two systems share one additional feature that most other systems lack: both are joined models that simultaneously learn the sentence-level gapping class and token-level gapping element labels.

We assume that this feature is relevant because it allows systems to minimize false positive examples for  the gap resolution and full annotation tasks. Since false positive examples receive a rather harsh score penalty, joint training could potentially offer a substantial score improvement for the whole system.

\section{Conclusion}
We have presented the AGRR-2019 gapping corpus for Russian. Our corpus contains 22.5k sentences, including 7.5k sentences with gapping and 15k relevant negative sentences. The corpus is multi-genre and social media texts form a quarter of it.

It should be noted that to the best of our knowledge no other publicly available corpus for any language contains a comparable number of gapping examples. We believe that theoretical studies may also benefit from this data.

We have developed an annotation scheme that identifies gapping elements - parts of the sentence most relevant for gapping resolution from the theoretical point of view (see analysis in section \ref{Gapping}). Our annotation scheme allows for successful solution of gapping resolution tasks by modifying standard sequence labeling techniques.

An important property of the AGRR-2019 corpus is that the systems trained on this corpus yield low number of false positives. Given the fact that gapping is a naturally rare phenomenon, this feature is extremely important.

While our corpus has some inherent limitations (see Section \ref{SynTagRus}), the evaluation of the top system on the SynTagRus gapping test set demonstrates that the AGRR-2019 corpus is not an artificial creation of Compreno parser, but rather covers a large subset of Russian language gapping (see Section \ref{Evaluation}).

We hope that the size and diversity of our corpus will provide researchers interested in gapping with a valuable source of information that could bring the community closer to resolving ellipsis.

The corpus described in this paper can be utilized to improve parsing quality, possibly not only for Russian but for other Slavic languages as well.

\section{Acknowledgments}
We are grateful to Alexey Bogdanov for his help with linguistic analysis and useful advice. We are thankful to Ekaterina Lyutikova for her insightful comments on the SynTagRus gapping test set. We thank the anonymous reviewers for their valuable feedback.

The work was partially supported by the GA UK grant 794417.

\bibliography{agrr}
\bibliographystyle{acl_natbib}

\end{document}